# Automatic Detection of Injection and Press Mold Parts on 2D Drawing Using Deep Neural Network

Junseok Lee[1], Jongwon Kim[1], Jumi Park[1], Seunghyeok Back[1], Seongho Bak[1],
Kyoobin Lee[1*]

[1] School of Integrated Technology (SIT), Gwangju Institute of Science and Technology (GIST),
Gwangju, 61005, Korea (Junseoklee@gm.gist.ac.kr, Jongwonkim@gm.gist.ac.kr, jumipark1126@gm.gist.ac.kr, shback@gist.ac.kr, bakseongho@gm.gist.ac.kr)

[1] School of Integrated Technology (SIT), Gwangju Institute of Science and Technology (GIST),
Gwangju, 61005, Korea (kyoobinlee@gist.ac.kr) * Corresponding author

**Abstract:** This paper proposes a method to automatically detect the key feature parts in a CAD of commercial TV and monitor using a deep neural network. We developed a deep learning pipeline that can detect the injection parts such as hook, boss, undercut and press parts such as DPS, Embo-Screwless, Embo-Burring, and EMBO in the 2D CAD drawing images. We first cropped the drawing to a specific size for the training efficiency of a deep neural network. Then, we use Cascade R-CNN to find the position of injection and press parts and use Resnet-50 to predict the orientation of the parts. Finally, we convert the position of the parts found through the cropped image to the position of the original image. As a result, we obtained detection accuracy of injection and press parts with 84.1% in AP (Average Precision), 91.2% in AR(Average Recall), 72.0% in AP, 87.0% in AR, and orientation accuracy of injection and press parts with 94.4% and 92.0%, which can facilitate the faster design in industrial product design

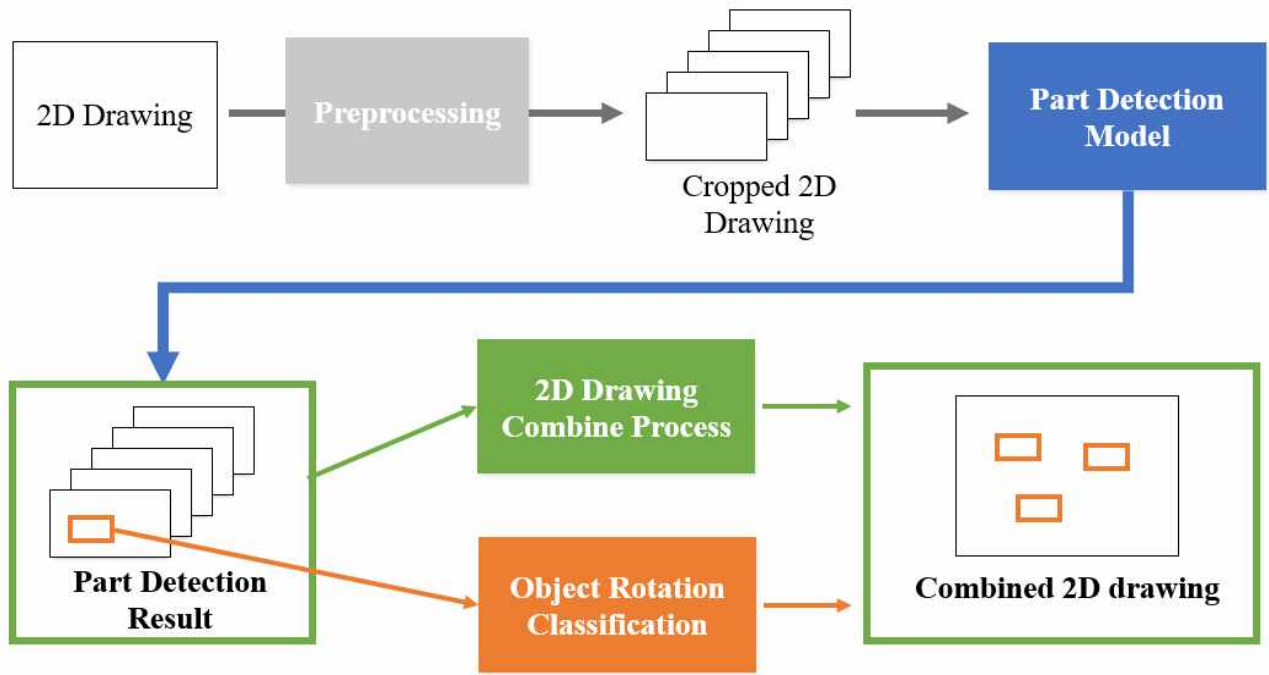

Fig. 1. The framework of the proposed method

## 1. INTRODUCTION

Recently, 2D injection and press mold are widespread manufacturing method in the industry [6]. Therefore, the demand for mold manufacturing is increasing and the limit of time is an issue. To manufacture mold parts, mold designers need to design molds that can print parts. Usually, TV or monitor chassis parts have flat, with several features aligned throughout the chassis.

At present, when designing a TV or monitor chassis, the shape of the mold modeling that the designer wants to place is all manually arranged by the designer. Due to the limitations of the current working method of manually placing features, mold design CAD/CAM design time reduction is difficult. The way in which all features are placed by designers depends only on the capabilities of mold designers. A feature recognition toolbox may be used to reduce dependence on worker competence, but it is not applicable to the current job due to the need to detect arbitrary features. Even with the help of a toolbox, there is a limit to the speed of a person's work. Therefore, an automation system is needed to handle many kinds of mold designs.

In this research, we would like to obtain the position and rotation angle of a particular part feature in the 2D drawing. Studies that detect specific symbols, shapes, and shapes in existing 2D drawings have been conducted in several directions using machine learning, graphical network use, and deep learning [5, 8, 10, 11]. In particular, the deep learning has shown outstanding results in the object detection and classification areas [4, 7, 12, 13].

We solved the problem through deep learning, using the Regions with Convolution Neural Network features (R-CNN)[9] method for object detection that identifies geometry, and the Convolution Neural Network (CNN) for rotation angle classification separately.

The features detected in the drawing are displayed in the form of a bounding box, with the position, type, rotation angle, and size of the bounding box being displayed as an annotation.

The results of this study can be applied to the improvement of the mold manufacturing process. Using the results of the study, it is possible to determine product appearance problems in the injection and trial of fine blanking parts by using the information of shape located in the mold design analysis. In the mold manufacture process, it is possible to decide whether to complete the targeted shape part milling or add electric discharge machining. In the mold assembly process, shape recognition technology can be used to create an assembly location map or be expected to assembly automation through robots.

## 2. METHOD

**2.1 Data preprocessing**

We propose an algorithm to find the location, type, and rotation angle of parts corresponding to the injection mold and press mold in the TV drawing. We use R-CNN (Region proposal + convolutional neural network) [9] to detect the location and type of parts in the image. Preprocessing is required for the 3D television CAD to enter the model's input because the R-CNN models analyze 2d images.

In the research, we convert 3D CAD created by 3D TV CAD production experts into 2D drawing images. There are 192 TV drawings, and the size of the drawing ranges from a minimum of 15000 x 8000 to a maximum of 30000 x 16000. Due to its large size, it is necessary to preprocess the large drawing image to use as input of the deep neural network.

First, we proceeded with annotation of the target objects present in the drawings for effective pretreatment. The location information of the target object is stored and organized in bounding box form. Next, cropped image used as input of the deep neural network so that the target object in the original drawing image was not cut as much as possible. Injection molds, 1333 x 800 was cropped to ensure sufficient features were present in the image and 2666 x 1600 was cropped to press molds.

There is not enough amount of data when we simply crop the drawing. To get enough data from limited sources, we applied a method to crop images by overlapping them by a certain percentage. Therefore, we slide the window of 1333x800 size when we crop the image in the whole drawing. This window moves from left to right so that it overlaps 90% with the previous position.

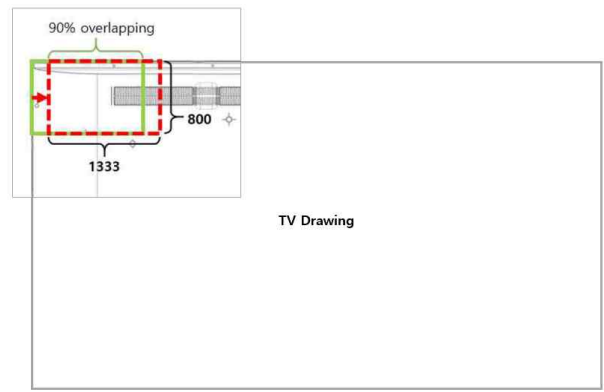

Fig. 2. Example of data preprocess

**2.2. Object detection**

In the process of mold design, a human manually places a 3D mold part at the position of the part existing in the 2D drawing. Therefore, it is necessary for a person to find the position and type of injection and press parts in the 2D drawing. This process needs a lot of time and causes fatigue because human do it manually. So we aimed to reduce design time by automating part detection in 2D drawing.

The object detection is to find out the position and class of an object in an image. We used a cropped image generated by data preprocessing as input to the object detection network. We use a detection network to locate injection and press parts and to classify the parts. The 2D drawing of injection mold has three parts : Hook, Boss, Undercut. The 2D drawing of press mold has four parts : DPS, Embo-screwless, Embo-Burring, EMBO.

We use Cascade R-CNN, which show high performance in object detection. The cascade R-CNN used a method of training by gradually increasing the Intersection of Union(IoU). The detector generated increasingly accurate proposals as it trains from low IoU to high IoU.

**2.3 Orientation prediction**

Based on the output of object detection from cropped batch, we classify rotation angle of parts. On the TV drawing, rotation of parts is contained in up to 4 cases (degree of 0,90,180, and 270). In the 7 classes of parts, "Boss", "Embo" and "Embo burring" does not have rotation because of its symmetric shape. On the other hand, "DPS" has only 2 rotations (0,90), and "Hook", "Undercut" and "Embo screwless" have 4 rotations. Examples of rotated part's images can be described by Figure 3.

We used separated two classification models that classify rotation of injection and press parts. In the other words, one model classifies 8 classes for injection parts (4 for "hook" and 4 for "undercut"), the other classifies 6 classes of press parts (2 for "DPS" and 4 for "Embo screwless").

The training data is acquired from output of object detection model which is described in 2.2. Also, we selected 1500 images for each rotation of each part for evaluation. We used ResNet 50[5] for classification, input images are resized by 224x224.

Fig. 3. Example of rotated parts

## 3. EXPERIMENTAL RESULTS

### 3.1 Experiment setting

We experimented that the Cascade RCNN [4] model on our preprocessed 2d drawing data. We conducted training and testing for two types of injection mold and press mold. The model was trained separately on injection parts and press parts. There are 122 drawings for injection parts. 92 drawings are used for training and 30 drawings are used for test. Similarly, 54 drawings are used as train set, and 16 drawings are used as test set for press parts detection. Model evaluation was done for both object detection and classification.

### 3.2 Object detection experiment

The detection result was evaluated by average precision (AP) and average recall (AR). The two evaluation methods are primarily used to evaluate object detection algorithms and have some differences. The precision refers to the proportion of the correct area of all detection results. On the contrary, the recall refers to the proportion of areas that need to be detected, which are properly detected. The evaluation value is calculated by varying the IoU threshold from 0.5 to 0.95, by 0.05 and averaging the values obtained. The precision value of the injection mold is 84.1%, the Recall value is 91.2%. The press mold's precision value is 72%, recall value is 87%. Both injection molds and press molds showed some difference, but they showed the results of detecting the target objects present in the drawing.

Table 1.

|  | AP@50:5:95 | AR |
|---|---|---|
| Injection | 84.1% | 91.2% |
| Press | 72% | 87% |

### 3.3 Orientation prediction experiment

The classification of object's rotation was evaluated using accuracy. The injection molds were performed for Hook, Undercut, and press molds were evaluated for DPS, EMBO. The reason for not evaluating the angles of the entire part was excluded because certain parts do not change their appearance even if they are symmetrical or rotated. Therefore, the evaluation was conducted only on parts that had changes when rotated. The angle classification accuracy of injection molds and press molds is 94.4% and 92%, respectively. The test result showed that the rotation angle was adjusted with a high probability.

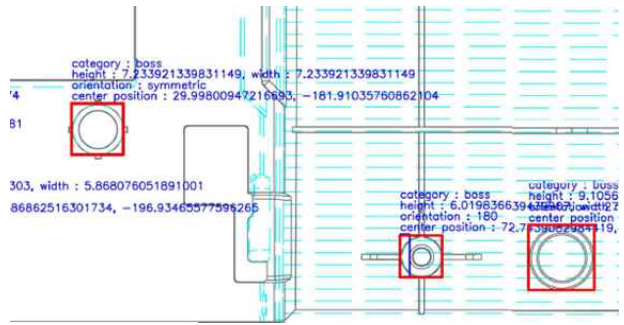

Fig. 4. Inference example of object detection result

## 4. CONCLUSION

In this paper, we present automatic detection network that detects injection and press parts and predicts orientation on 2D drawing. We used the 2D drawing which is a minimum size of 15000x8000 and a maximum size of 30000x16000. We have created a process for detection on images that are much larger than those used in typical object detection. The results of this study show performance of injection and press parts in 2D drawing. We determine that our automatic detection network can be used sufficiently for mold part recognition for automated mold design.

## ACKNOWLEDGEMENT

This paper is a research conducted with the support of

## REFERENCES

[1] Z. Cai and N. Vasconcelos, "Cascade R-CNN: High Quality Object Detection and Instance Segmentation," *IEEE Transactions on Pattern Analysis and Machine Intelligence*, vol. 43, no. 5, pp. 1483–1498, 2021.

[2] K. He, X. Zhang, S. Ren, and J. Sun, "Deep Residual Learning for Image Recognition," *2016 IEEE Conference on Computer Vision and Pattern Recognition (CVPR)*, 2016.

[3] Kuo, Chil-Chyuan, and Xin-Yi Yang. "Optimization of direct metal printing process parameters for plastic injection mold with both gas permeability and mechanical properties using design of experiments approach." *The International Journal of Advanced Manufacturing Technology* 109.5 (2020): 1219-1235.

[4] He, Kaiming, et al. "Mask r-cnn." *Proceedings of the IEEE international conference on computer vision*. 2017.

[5] Girshick, Ross. "Fast r-cnn." *Proceedings of the IEEE international conference on computer vision*. 2015.

[6] Girshick, Ross, et al. "Rich feature hierarchies for accurate object detection and semantic segmentation." *Proceedings of the IEEE conference on computer vision and pattern recognition*. 2014.

[7] Simonyan, Karen, and Andrew Zisserman. "Very deep convolutional networks for large-scale image recognition." *arXiv preprint arXiv:1409.1556* (2014).

[8] Redmon, Joseph, et al. "You only look once: Unified, real-time object detection." *Proceedings of the IEEE conference on computer vision and pattern recognition*. 2016.

[9] Carion, Nicolas, et al. "End-to-end object detection with transformers." *European Conference on Computer Vision*. Springer, Cham, 2020.

[10] Tan, Mingxing, Ruoming Pang, and Quoc V. Le. "Efficientdet: Scalable and efficient object detection." *Proceedings of the IEEE/CVF conference on computer vision and pattern recognition*. 2020.